\documentclass[pdflatex,sn-mathphys-num]{sn-jnl}% Math and Physical Sciences Numbered Reference Style
\usepackage{graphicx}%
\usepackage{multirow}%
\usepackage{amsmath,amssymb,amsfonts}%
\usepackage{amsthm}%
\usepackage{mathrsfs}%
\usepackage[title]{appendix}%
\usepackage{xcolor}%
\usepackage{textcomp}%
\usepackage{manyfoot}%
\usepackage{booktabs}%
\usepackage{algorithm}%
\usepackage{algorithmicx}%
\usepackage{algpseudocode}%
\usepackage{listings}%
\usepackage{comment}
\theoremstyle{thmstyleone}%
\theoremstyle{thmstyletwo}%

\theoremstyle{thmstylethree}%

\begin{document}

\title[Article Title]{Effects of personality steering on cooperative behavior in Large Language Model agents}


\author[1]{\fnm{Mizuki} \sur{Sakai}}\email{sakai.mizuki.20@shizuoka.ac.jp}

\author[1]{\fnm{Mizuki} \sur{Yokoyama}}\email{yokoyama.mizuki.24@shizuoka.ac.jp}

\author[2]{\fnm{Wakaba} \sur{Tateishi}}\email{2wakaba.tateishi@gmail.com}

\author*[1]{\fnm{Genki} \sur{Ichinose}}\email{ichinose.genki@shizuoka.ac.jp}

\affil*[1]{\orgdiv{Department of Mathematical and Systems Engineering}, \orgname{Shizuoka University}, \orgaddress{\street{3-5-1}, \city{Hamamatsu}, \postcode{4328561}, \state{Shizuoka}, \country{Japan}}}

\affil[2]{\orgdiv{Faculty of Business}, \orgname{Hokkaido Musashi Women's University}, \orgaddress{\street{Kita 22-jo Nishi 13-chome}, \city{Sapporo}, \postcode{0010022}, \state{Hokkaido}, \country{Japan}}}


\abstract{Large language models (LLMs) are increasingly used as autonomous agents in strategic and social interactions.
Although recent studies suggest that assigning personality traits to LLMs can influence their behavior, how personality steering affects cooperation under controlled conditions remains unclear.
In this study, we examine the effects of personality steering on cooperative behavior in LLM agents using repeated Prisoner’s Dilemma games.
Based on the Big Five framework, we first measure basic personality scores of three models, GPT-3.5-turbo, GPT-4o, and GPT-5, using the Big Five Inventory.
We then compare behavior under baseline and personality-informed conditions, and further analyze the effects of independently manipulating each personality dimension to extreme values.
Our results show that agreeableness is the dominant factor promoting cooperation across all models, while other personality traits have limited impact.
Explicit personality information increases cooperation but can also raise vulnerability to exploitation, particularly in earlier-generation models.
In contrast, later-generation models exhibit more selective cooperation.
These findings indicate that personality steering acts as a behavioral bias rather than a deterministic control mechanism.}




\maketitle

\section{Introduction}\label{sec1}

Multi-agent systems have long been studied and utilized to make decisions on behalf of humans.
Traditional multi-agent systems were designed to achieve rational or cooperative behavior under specific environments, based on limited information-processing capabilities and predefined rules.
Typical applications include price negotiations and contract adjustments, where multiple agents seek agreement while pursuing conflicting interests \cite{Shoham2009}; resource allocation scenarios, where competition and cooperation coexist over limited resources \cite{Perolat2017}; and marketplaces or auctions, where seller and buyer agents strategically propose prices \cite{Wellman2016,Shoham2009}.
In recent years, the emergence of Large Language Models (LLMs) has substantially expanded the decision-making capabilities of agents.
By employing LLMs as agents, flexible interpretation of information and reasoning through natural language become possible, enabling decision-making even in complex and context-dependent social situations that were previously difficult to handle.
As a result, LLM-based multi-agent systems have demonstrated new potential in social interactions involving multiple stakeholders and in settings that require human-like judgment \cite{Park2023,Newsham2025,Zeng2025,Akata2025}.

However, this shift in agent capabilities has also introduced new challenges.
Interactions among LLM-based agents can be unpredictable \cite{Park2023} and may involve the risk of escalating unintended conflicts \cite{Rivera2024}.
In strategic environments, optimal decision-making requires not only the maximization of immediate rewards but also careful consideration of relationships with other agents and the formation of long-term cooperative relationships.
Such instability highlights the need for new approaches to guide and control the behavior and decision-making of LLMs in strategic settings.
As a potential solution to this challenge, the psychological framework of personality has attracted increasing attention.
In personality psychology, various theories have been proposed to explain individual behavioral patterns, among which the Big Five Personality Traits \cite{John1991} are widely accepted as a theoretical framework that describes personality along five major dimensions: (1) Extraversion, which reflects sociability, activity level, and the tendency to seek stimulation; (2) Agreeableness, which captures empathy, cooperativeness, and tolerance toward others; (3) Conscientiousness, which represents responsibility, organization, and motivation toward goal achievement; (4) Neuroticism, which reflects emotional instability and sensitivity to stress; and (5) Openness to Experience, which captures openness to new experiences and ideas, as well as creativity.
In human psychological research, individuals with high agreeableness tend to exhibit stronger cooperative behavior, while also being more vulnerable to exploitation \cite{Thielmann2020}.
Moreover, in strategic situations such as repeated Prisoner's Dilemma games, individuals with higher agreeableness are more likely to cooperate in the early stages of interaction \cite{Kagel2014}. 

At the same time, because the Big Five framework provides well-established quantitative measurement methods, it is also well suited as a basis for analyzing personality traits in LLMs, suggesting that such traits may influence decision-making in these models.
Indeed, a previous study has reported that increasing traits such as agreeableness and conscientiousness in LLM agents promotes cooperative behavior, while simultaneously increasing vulnerability to exploitation by other agents, leading to lower overall payoffs-patterns that closely resemble those observed in humans \cite{Ong2025}.
However, this prior study has several important limitations.
First, it typically begins by assigning or manipulating personality traits in LLMs without first quantitatively measuring the basic personality scores inherently exhibited by the models, making it difficult to assess how personality interventions alter behavior relative to a model’s original tendencies.
Second, because personality traits are often not specified in quantitative terms, the strength of personality manipulations cannot be systematically compared, which limits reproducibility and hinders meaningful comparisons across studies.

To address these limitations, we investigate the relationship between personality traits and cooperative behavior in LLM agents under quantitatively controlled conditions using the Big Five Personality Traits framework.
Specifically, we examine three objectives.
First, we quantitatively measure the basic personality scores inherently exhibited by different LLMs based on the Big Five framework, thereby clarifying model-specific characteristics and differences across model generations.
Second, we examine how LLM agents behave in strategic settings such as repeated Prisoner’s Dilemma games when their measured personality traits are explicitly provided through prompts.
Third, we analyze how cooperative behavior changes when each personality dimension is independently manipulated to extreme values while holding the remaining dimensions constant.
Through these analyses, we aim to clarify how personality steering influences cooperative behavior in LLM agents and to provide a controlled framework for understanding the role of personality traits in LLM decision-making.

\section{Methods}\label{sec11}

In this study, we investigate the relationship between personality traits and cooperative behavior in LLM agents using three widely used models developed by OpenAI: GPT-3.5-turbo, GPT-4o, and GPT-5.
We conduct the experiments in three stages.
First, we quantitatively measure the basic personality scores of each model based on the Big Five Personality Traits using the Big Five Inventory (BFI-44) (Experiment 1).
Second, we examine how LLM agents behave in strategic settings by having them play repeated Prisoner’s Dilemma games under two conditions: a baseline condition without explicit personality information and a condition in which the measured personality traits obtained in Experiment 1 are explicitly provided via prompts (Experiment 2).
Third, we analyze the effects of personality manipulation by independently setting each Big Five trait to extreme values while keeping the remaining traits constant, in order to assess which personality dimensions most strongly influence cooperative behavior (Experiment 3).

\subsection{Experiment 1: Measuring basic personality traits using the Big Five}

In Experiment 1, we first measure the personality traits of each model based on the Big Five Personality Traits framework \cite{John1991}.
%The Big Five model is a widely accepted theory in personality psychology that describes personality along five major dimensions: (1) Extraversion, which reflects sociability, activity level, and the tendency to seek stimulation; (2) Agreeableness, which captures empathy, cooperativeness, and tolerance toward others; (3) Conscientiousness, which represents responsibility, organization, and motivation toward goal achievement; (4) Neuroticism, which reflects emotional instability and sensitivity to stress; and (5) Openness to Experience, which captures openness to new experiences and ideas, as well as creativity.
To quantitatively assess these personality traits in LLMs, we employ the Big Five Inventory (BFI), one of the most widely used psychometric instruments for measuring personality traits \cite{John1991}.
The BFI is a self-report questionnaire developed based on the Big Five theory and has been validated across diverse cultures and languages, demonstrating high reliability and validity.
In this study, we use the BFI-44, a shortened version consisting of 44 items, with 8–10 items assigned to each personality dimension.
Each item is rated on a five-point Likert scale ranging from 1 (``strongly disagree") to 5 (``strongly agree").

For each model, the 44 questionnaire items are presented sequentially, with instructions prompting the model to respond to each item using a numerical value from 1 to 5 (see Appendix A for the full prompt).
We adopt the BFI prompt developed by Jiang et al. \cite{Jiang2024}.
If a response does not conform to the required format (e.g., contains non-numeric characters), the question is presented again until a valid numerical response is obtained.
After collecting responses to all items, scores for each personality dimension are computed following the standard BFI scoring procedure.
Because LLM outputs involve stochasticity, we repeat the personality measurement 20 times for each model and use the average scores across the 20 runs to improve score stability.
Through this procedure, we obtain quantitative and reproducible basic personality scores for each LLM.

\subsection{Experiment 2: Behavioral differences with explicit personality information}

In Experiment 2, we examine how LLM agents behave in strategic environments by having them play repeated Prisoner's Dilemma (RPD) games.
We consider two experimental conditions.
In the baseline condition, no explicit personality information is provided to the LLMs.
In the personality-informed condition, the personality traits measured in Experiment 1 are explicitly provided to the LLMs via prompts.
By comparing these two conditions, we assess whether and how explicitly supplied personality information influence cooperative behavior.

The Prisoner’s Dilemma is a canonical game-theoretic model that captures social dilemmas arising from the tension between cooperation and defection.
In each round, two players simultaneously choose either to cooperate (C) or defect (D).
Payoffs are determined by the joint action of both players according to a payoff matrix satisfying $T>R>P>S$ , where $T$ denotes the temptation to defect, $R$ the reward for mutual cooperation, $P$ the punishment for mutual defection, and $S$ the payoff received by a cooperator when the opponent defects.
In this study, we use the standard payoff values $T=5$, $R=3$, $P=1$, and $S=0$.

In the RPD game, the same pair of players interacts repeatedly, and it is assumed that each player can observe and remember past actions and outcomes. This repeated structure allows players to condition their current decisions on previous interactions, thereby enabling the emergence of cooperative behavior. Both theoretical \cite{Friedman1971,Kreps1982,Axelrod1981,Axelrod1984,Nowak1992,Nowak1993,Kandori1992} and experimental \cite{Selten1986,Andreoni1993,Bo2005,Camera2009,Wang2017} studies have demonstrated that cooperation can be sustained in the RPD games under appropriate conditions.

We first conduct the RPD experiments under the baseline condition to characterize the intrinsic cooperative tendencies of each LLM.
In this condition, the prompt include only the rules and objectives of the RPD game and does not contain any information related to personality traits.
The prompt is constructed based on that proposed by Fontana et al. \cite{Fontana2024}.
The full prompt details are provided in Appendix B.
Each LLM agent play 100 independent trials, each consisting of 10 rounds of the RPD game.
Trials are independent, and outcomes from previous trials do not affect subsequent trials.

In each trial, the LLM agent play against a set of fixed opponent strategies that are hard-coded and commonly used in studies of the RPD game.
These strategies include: Always Cooperate (ALLC), which selects cooperation in every round; Always Defect (ALLD), which selects defection in every round; Random (RANDOM), which randomly chooses between cooperation and defection in each round; Tit-for-Tat (TFT), which cooperates in the first round and then replicates the opponent’s previous action; and Grim Trigger (GRIM), which cooperates initially but switches permanently to defection after the opponent defects once.
Together, these strategies cover a broad range of cooperative and non-cooperative behaviors.

In the personality-informed condition, we repeat the same RPD experiments while explicitly providing each LLM with its measured Big Five personality scores obtained in Experiment 1. The prompt specifies the personality traits in the form: ``Your personality traits are as follows: Openness (O), Conscientiousness (C), Extraversion (E), Agreeableness (A), and Neuroticism (N)," followed by the corresponding numerical scores (see Appendix C for prompt details).
Under this condition, the LLM is instructed to make decisions while being aware of its own personality traits.

For each experimental condition, we compute two evaluation metrics.
The average cooperation rate is defined as the proportion of cooperative choices across all rounds, indicating the degree of cooperativeness exhibited by the LLM.
The average cumulative payoff is defined as the mean total payoff accumulated across all rounds, reflecting the overall performance of the LLM.
By comparing these metrics between the baseline and personality-informed conditions, we evaluate the behavioral impact of explicitly providing personality information to LLM agents.

\subsection{Experiment 3: Effects of extreme personality manipulation}

In Experiment 3, we investigate how artificial manipulation of personality traits affects the behavior of LLM agents by modifying the Big Five personality scores obtained in Experiment 1.
Specifically, we independently set one personality dimension, Agreeableness, Extraversion, Conscientiousness, Neuroticism, or Openness, to an extreme value of either 1 or 5, while keeping the remaining four dimensions fixed at their measured average values.
This procedure allows us to isolate the individual contribution of each personality dimension to cooperative behavior.

In each experimental condition, only one personality trait is manipulated.
For example, when manipulating agreeableness, the agreeableness score is set to either 1 (minimum) or 5 (maximum), while the scores for openness, conscientiousness, extraversion, and neuroticism remain unchanged from the values measured in Experiment 1.
By applying this manipulation independently to each of the five personality dimensions and considering both extreme values, we obtain a total of ten experimental conditions (five dimensions $\times$ two values).

For each condition, the modified personality scores are explicitly provided to the LLM agents via prompts in the same format as in the personality-informed condition of Experiment 2.
The agents then play the RPD games under identical settings, including the same opponent strategies, number of rounds, and number of trials.
Using the same evaluation metrics as in Experiment 2, the average cooperation rate and the average cumulative payoff, we assess how extreme manipulation of individual personality traits influence cooperative behavior.

\subsection{LLM settings}
In this study, we use three large language models developed by OpenAI: GPT-3.5-turbo, GPT-4o, and GPT-5.
These models represent successive generations of LLMs, with larger model numbers corresponding to more recent releases.

For GPT-3.5-turbo and GPT-4o, we set the temperature parameter to 0.7.
The temperature controls the degree of randomness in model outputs, with lower values producing more deterministic responses and higher values increasing stochasticity.
The value of 0.7 is chosen to balance response diversity and consistency, and is consistent with the setting used in prior work \cite{Fontana2024}.

For GPT-5, we conduct experiments with the reasoning effort set to minimal and the verbosity set to low.
The reasoning effort parameter controls the amount of computational effort allocated to the model’s internal reasoning process; setting it to minimal simplifies the reasoning process and improves response speed.
The verbosity parameter controls the level of detail in the model’s output; setting it to low ensures that only the minimum necessary information is produced.

\section{Results}\label{sec2}

\subsection{BFI scores}
\begin{figure}[htbp]
    \centering
    \includegraphics[width=\columnwidth]{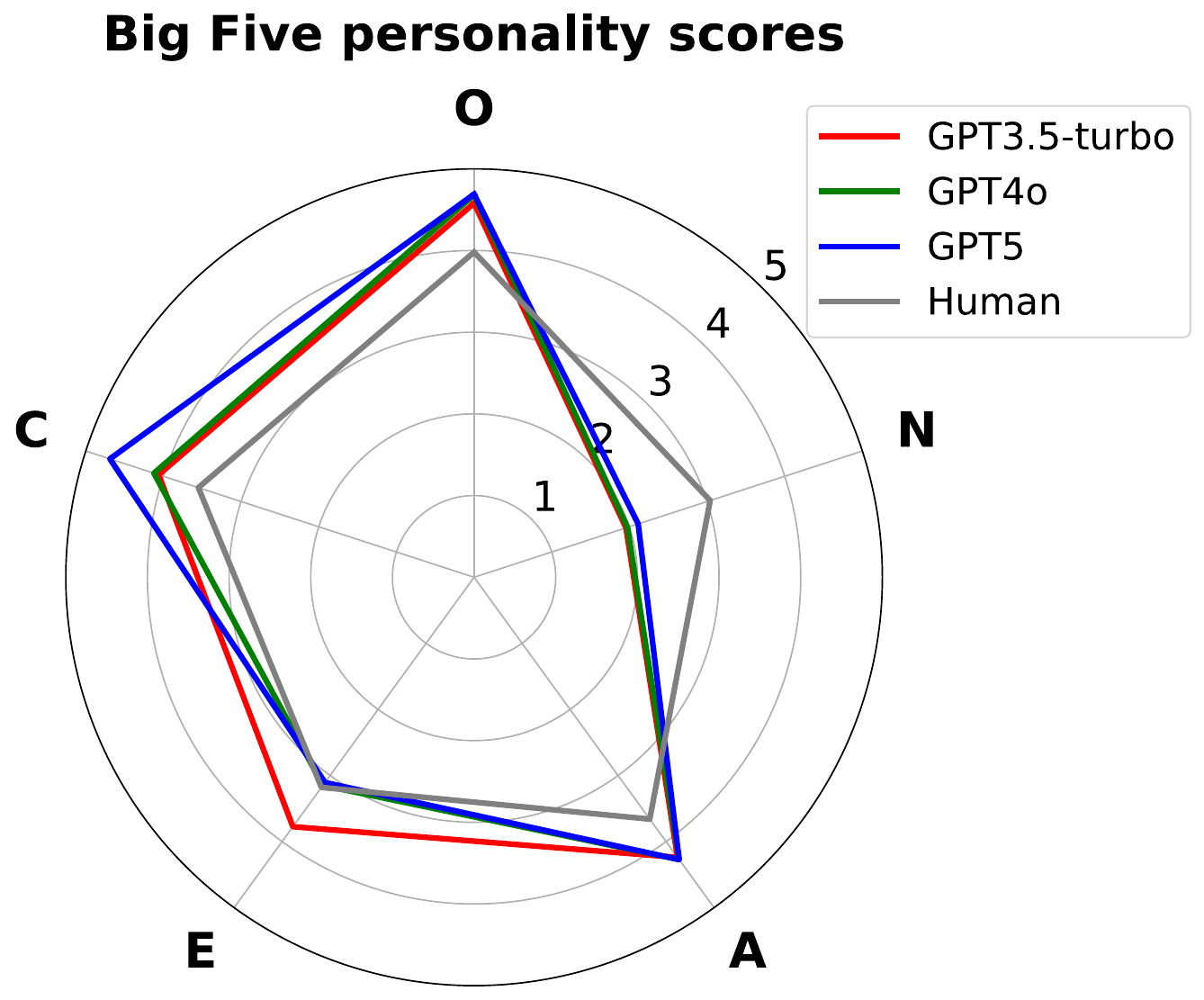}
    \caption{Radar plot of the average Big Five personality scores for GPT-3.5-turbo, GPT-4o, GPT-5, and humans. Scores for the models are averaged over 20 repeated measurements.
}
    \label{fig:bfi}
\end{figure}

We first report the results of Experiment 1, in which the personality traits of each LLM were measured using the Big Five Inventory.
Table \ref{tab:bfi_scores} presents the mean scores and standard deviations obtained from 20 repeated measurements for each model.
For comparison, the table also includes human data reported by Srivastava et al. \cite{Srivastava2003}, where the original POMP scores were rescaled to the same 1–5 scale used in this study.
Figure \ref{fig:bfi} provides a radar-plot visualization of the average Big Five personality scores shown in Table \ref{tab:bfi_scores}.

\begin{table*}[h]
    \centering
    \resizebox{\textwidth}{!}{%
    \begin{tabular}{c|cccc}
    \hline
    \textbf{Personality trait} & \textbf{GPT-3.5-turbo} & \textbf{GPT-4o} & \textbf{GPT-5} & \textbf{Human} \\
    \hline
    Openness (O) & 4.58 (0.12) & 4.68 (0.25) & 4.69 (0.17) & 3.98 (0.66) \\
    Conscientiousness (C) & 4.06 (0.12) & 4.12 (0.32) & 4.69 (0.16) & 3.55 (0.73) \\
    Extraversion (E) & 3.78 (0.11) & 3.15 (0.28) & 3.10 (0.37) & 3.18 (0.90) \\
    Agreeableness (A) & 4.24 (0.13) & 4.27 (0.32) & 4.27 (0.18) & 3.66 (0.72) \\
    Neuroticism (N) & 1.96 (0.20) & 1.98 (0.39) & 2.11 (0.21) & 3.04 (0.88) \\
    \hline
    \end{tabular}%
    }
    \caption{Mean BFI scores and standard deviations (in parentheses) for each model.}
    \label{tab:bfi_scores}
\end{table*}

Across all models, the 20 measurements yielded highly consistent scores, with relatively small standard deviations.
Specifically, the standard deviations ranged from 0.11 to 0.20 for GPT-3.5-turbo, from 0.25 to 0.39 for GPT-4o, and from 0.16 to 0.37 for GPT-5.
These values are substantially smaller than those observed in humans (0.66–0.90), indicating that each LLM exhibits a stable and internally consistent personality score.
This consistency suggests that the measured personality traits reliably characterize each model.

In contrast, the human scores cluster around the mid-range of the scale, with larger standard deviations, reflecting substantial inter-individual variability in human personality traits.
Compared to human averages, all LLMs exhibit higher scores in openness, conscientiousness, and agreeableness, as well as lower scores in neuroticism.
This pattern suggests that LLMs tend to exhibit personality scores that are systematically shifted toward higher scores in certain traits compared to humans, potentially reflecting biases toward socially desirable response patterns present in their training data.

Next, we examine the patterns observed on individual personality dimensions.
Overall, all models exhibited broadly similar tendencies, characterized by relatively high scores in openness, conscientiousness, and agreeableness, and low scores in neuroticism. This consistency across models suggests that contemporary LLMs share common personality scores as measured by the BFI framework.
A particularly notable result is observed for conscientiousness. While GPT-3.5-turbo and GPT-4o show relatively similar mean scores for conscientiousness (4.06 and 4.12), GPT-5 exhibits a higher score (4.69).
One possible explanation for this difference is the technical improvements reported for GPT-5, including a significant reduction in hallucinated responses and enhanced reasoning consistency compared to earlier models \cite{Openai2025}.
Such improvements may contribute to more reliable, goal-oriented, and internally consistent responses, which are reflected in higher conscientiousness scores.

\subsection{Comparison between baseline and personality-informed conditions}
\begin{figure}[htbp]
    \centering
    \includegraphics[width=\columnwidth]{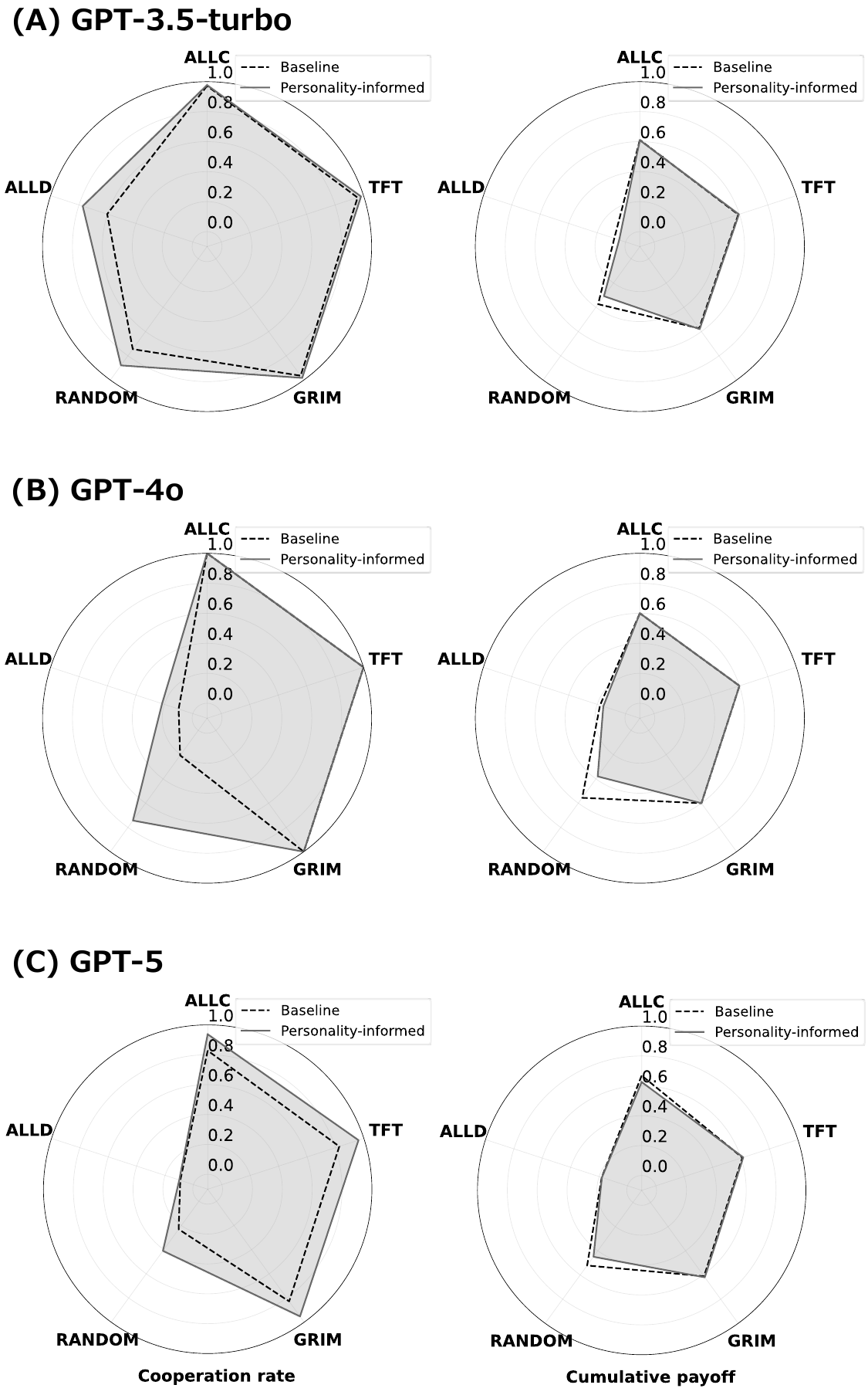}
    \caption{Average cooperation rates (left) and average cumulative payoffs (right) under the baseline condition (dashed) and the personality-informed condition (solid): (A) GPT-3.5-turbo, (B) GPT-4o, and (C) GPT-5.}
    \label{fig:baseline_vs_measured}
\end{figure}

We next compare the results of the repeated Prisoner’s Dilemma experiments under the baseline condition and the personality-informed condition.
Figure \ref{fig:baseline_vs_measured} shows the average cooperation rates and average cumulative payoffs for each model, where dashed lines indicate the baseline condition and solid lines indicate the personality-informed condition.
The average cumulative payoff is normalized to range between 0 and 1 for ease of comparison across models.

Across all models, the personality-informed condition yielded cooperation rates that were comparable to or higher than those observed in the baseline condition.
This pattern is consistent with the high agreeableness scores measured in Experiment 1 and suggests that explicitly providing personality information, particularly high agreeableness, promotes cooperative behavior in LLM agents.
These results indicate that personality steering can systematically influence agent behavior in strategic interactions.

For GPT-3.5-turbo (Fig.~\ref{fig:baseline_vs_measured}A), cooperation rates against non-cooperative strategies such as ALLD and RANDOM were relatively high even in the baseline condition, compared to GPT-4o (Fig.~\ref{fig:baseline_vs_measured}B) and GPT-5 (Fig.~\ref{fig:baseline_vs_measured}C).
Under the personality-informed condition, cooperation rates against these strategies increased substantially, while average cumulative payoffs decreased slightly.
This result indicates that although explicit personality information enhances cooperation, it also increases vulnerability to exploitation by defecting or unpredictable opponents.
In contrast, cooperation rates and payoffs against cooperative or reciprocal strategies, ALLC, GRIM TRIGGER, and TFT, remained high and largely unchanged across both conditions in GPT-3.5-turbo.

GPT-4o exhibited lower cooperation rates against ALLD and RANDOM strategies in the baseline condition, indicating more cautious behavior toward non-cooperative opponents (Fig.~\ref{fig:baseline_vs_measured}B).
However, under the personality-informed condition, cooperation rates against these strategies increased, accompanied by a reduction in cumulative payoffs.
Notably, the increase in cooperation was more pronounced for RANDOM than for ALLD, suggesting that GPT-4o distinguishes between purely defecting strategies and those that allow for potential reciprocal cooperation.

For GPT-5 (Fig.~\ref{fig:baseline_vs_measured}C), baseline behavior against ALLD was broadly similar to that of GPT-4o, whereas cooperation against RANDOM was substantially lower.
In addition, cooperation rates against ALLC, GRIM TRIGGER, and TFT were lower than those observed for GPT-4o.
By closely examining the game action histories, we found that GPT-5 frequently defected in the final round while maintaining cooperation in earlier rounds, a pattern consistent with end-game optimization in finitely repeated games.
Notably, such payoff-maximizing and seemingly self-interested behavior is consistent with prior findings showing that models with explicit reasoning capabilities tend to exhibit more self-interested decision-making compared to models without explicit reasoning processes \cite{Li2025}.
Under the personality-informed condition, cooperation rates against ALLC, GRIM TRIGGER, and TFT approached 100\%.
Cooperation rates against ALLD remained largely unchanged, and cooperation against RANDOM increased only modestly, indicating that GPT-5 limited cooperation with exploitative opponents even when personality information promoted cooperativeness.

Comparing the impact of personality steering across models reveals clear generational differences.
For GPT-3.5-turbo and GPT-4o, explicit personality information leads to higher cooperation rates but lower cumulative payoffs, reflecting a classic trade-off between cooperation and susceptibility to exploitation.
In contrast, GPT-5 exhibits a more selective response: while cooperation with reciprocal strategies increases, cooperation with non-cooperative strategies remains constrained.
This pattern suggests that newer models can incorporate personality steering while maintaining strategic robustness.

\subsection{Effects of extreme personality manipulation}
\begin{figure}[htbp]
    \centering
    \includegraphics[width=\columnwidth]{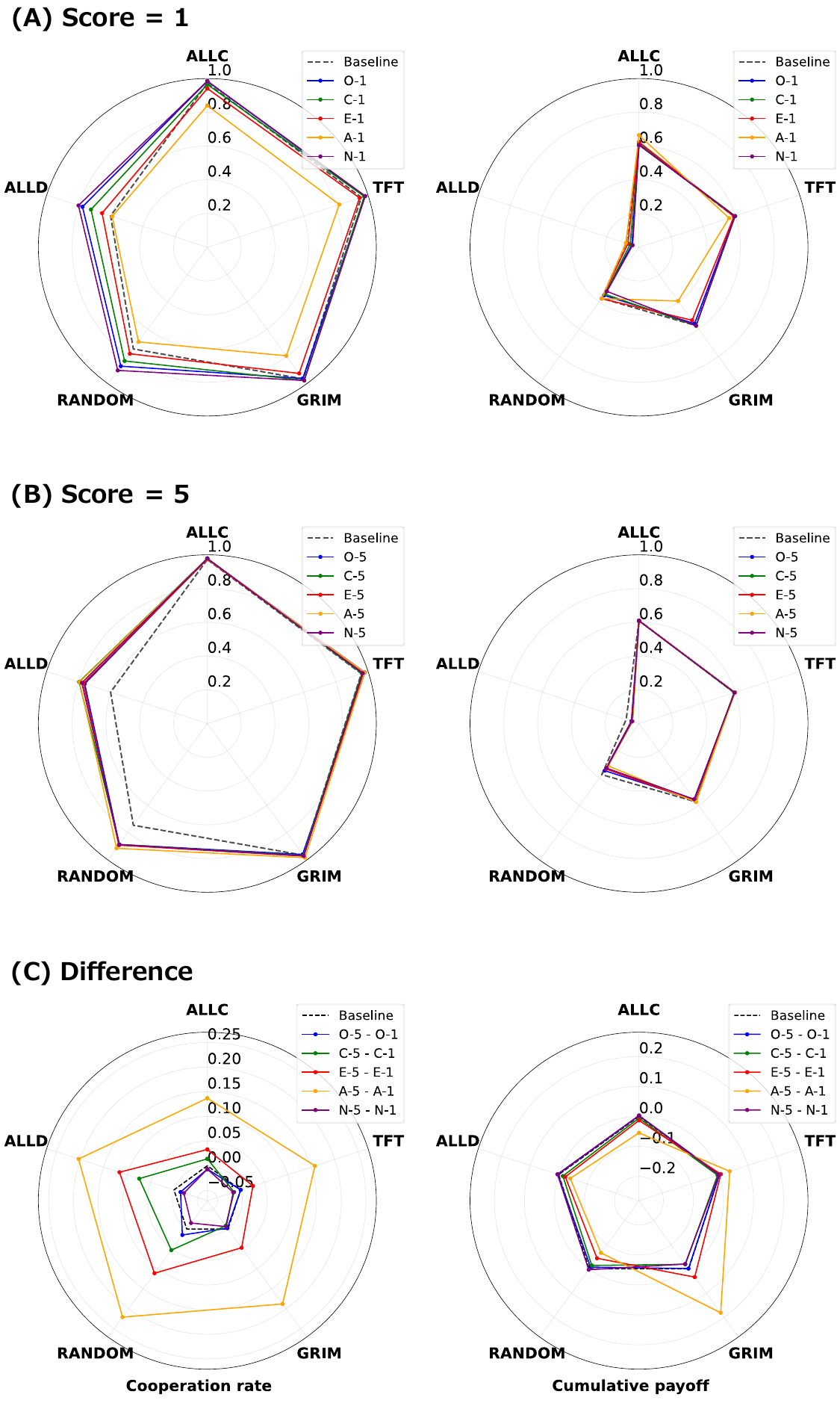}
    \caption{Effects of extreme personality-score manipulations for GPT-3.5-turbo. Left panels show cooperation rates and right panels show cumulative payoffs. (A) Manipulated personality dimension set to score = 1. (B) Manipulated personality dimension set to score = 5. (C) Differences between score = 5 and score = 1 ($5 - 1$).}
    \label{fig:model_35turbo}
\end{figure}

\begin{figure}[htbp]
    \centering
    \includegraphics[width=\columnwidth]{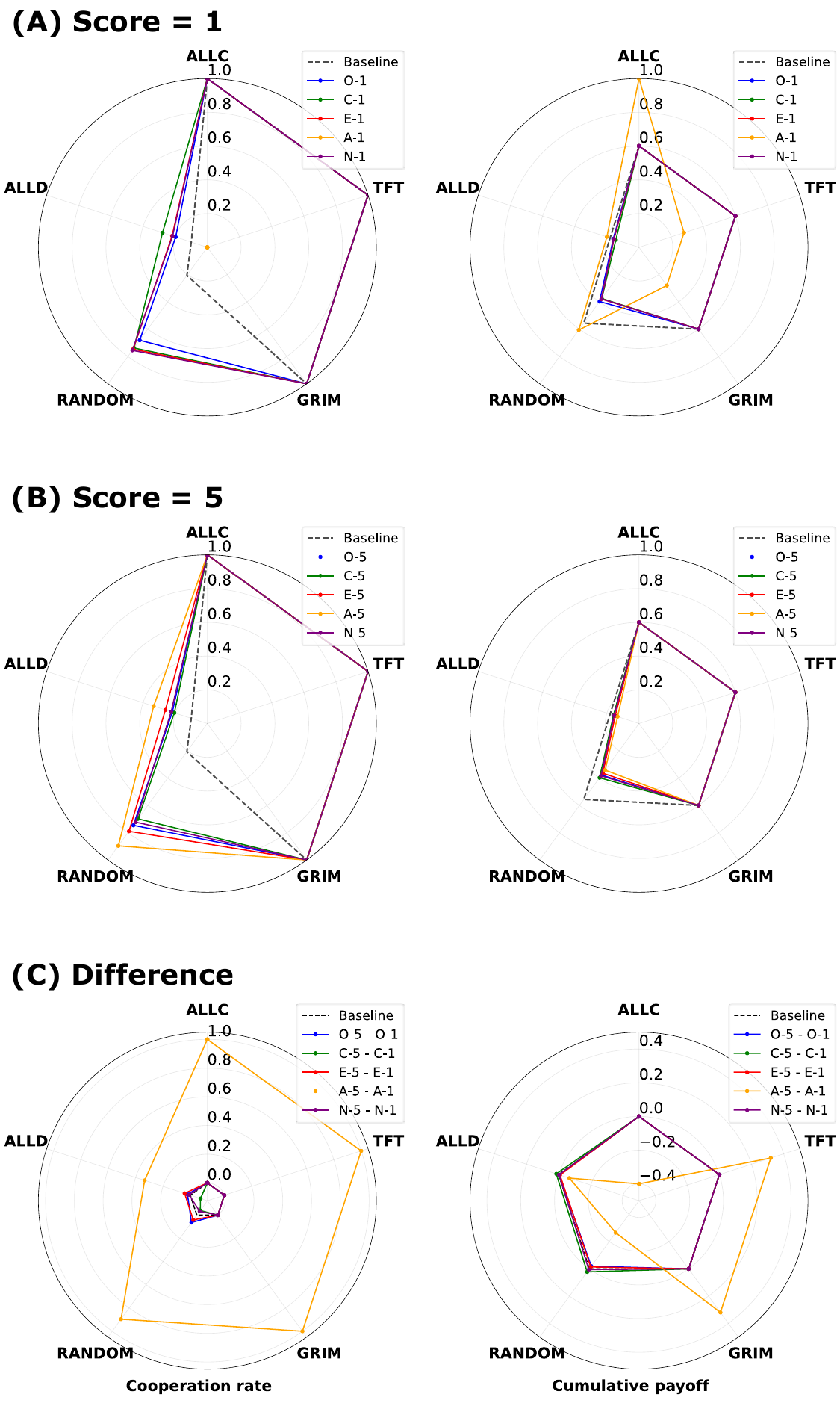}
    \caption{Effects of extreme personality-score manipulations for GPT-4o. Panel layout and notation are the same as in Fig.~\ref{fig:model_35turbo}.}
    \label{fig:model_4o}
\end{figure}

\begin{figure}[htbp]
    \centering
    \includegraphics[width=\columnwidth]{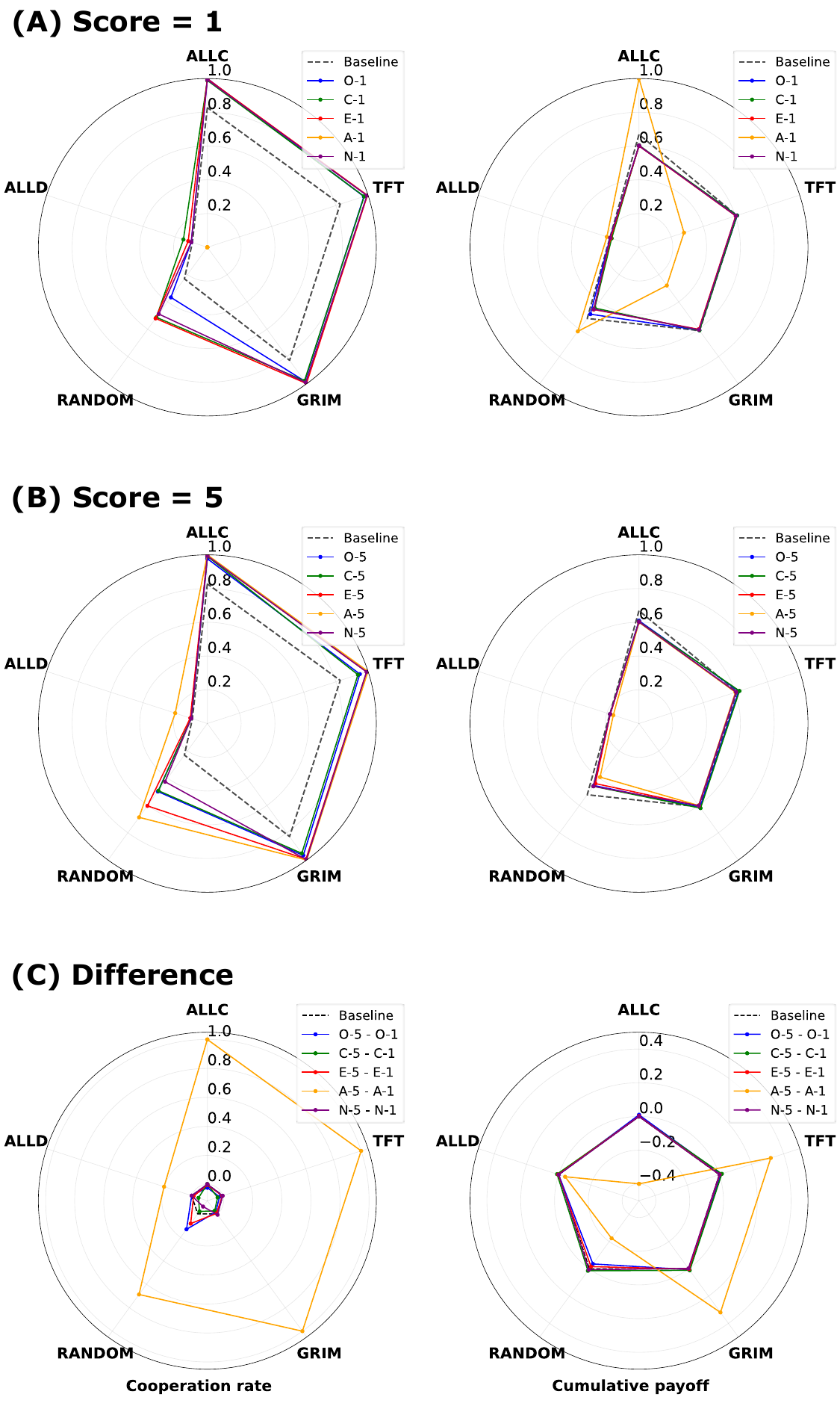}
    \caption{Effects of extreme personality-score manipulations for GPT-5. Panel layout and notation are the same as in Fig.~\ref{fig:model_35turbo}.}
    \label{fig:model_5}
\end{figure}

Finally, we report the results of Experiment 3, in which we artificially manipulated individual Big Five personality traits to extreme values.
Using the average BFI scores obtained in Experiment 1 as a reference, we independently replaced one personality dimension, openness, conscientiousness, extraversion, agreeableness, or neuroticism, with an extreme value of either 1 or 5, while keeping the remaining dimensions fixed.
The RPD experiments were then conducted under the same settings as in Experiment 2.
Figures \ref{fig:model_35turbo}, \ref{fig:model_4o}, and \ref{fig:model_5} present the average cooperation rates, average cumulative payoffs, and their differences for GPT-3.5-turbo, GPT-4o, and GPT-5, respectively.
In each figure, panels (A), (B), and (C) correspond to setting the manipulated personality dimension to score = 1, score = 5, and their difference ($5 - 1$), respectively.

For GPT-3.5-turbo (Fig.~\ref{fig:model_35turbo}), manipulation of agreeableness had the strongest impact on cooperative behavior, followed by extraversion.
When agreeableness was set to 1 (A-1) (Fig.~\ref{fig:model_35turbo}A), cooperation rates were consistently lower than those observed under other personality manipulations across all opponent strategies, resulting in a large negative difference in average cooperation.
In contrast, when agreeableness was set to 5 (A-5) (Fig.~\ref{fig:model_35turbo}B), cooperation rates were comparable to those observed under other personality conditions, yielding relatively small differences.

A similar but weaker pattern is observed for extraversion.
Under the E-1 condition (Fig.~\ref{fig:model_35turbo}A), cooperation rates against ALLD and RANDOM strategies were slightly lower than those under other personality manipulations, which contributed to observable differences in average cooperation.
When extraversion was set to 5 (E-5) (Fig.~\ref{fig:model_35turbo}B), however, cooperation rates showed little deviation from other conditions.

For GPT-4o (Fig.~\ref{fig:model_4o}), agreeableness again emerged as the dominant factor influencing cooperative behavior, whereas manipulation of other personality traits had minimal effects.
Under the A-1 condition (Fig.~\ref{fig:model_4o}A), cooperation rates droped to 0\% across all opponent strategies, indicating consistently defecting behavior.
Under the A-5 condition (Fig.~\ref{fig:model_4o}B), average cooperation rates increased to levels comparable to or higher than those observed under other personality manipulations.
However, cooperation rates against ALLD and RANDOM strategies were highest under A-5, resulting in increased exploitation and, consequently, the lowest average cumulative payoff among all personality conditions.

For GPT-5 (Fig.~\ref{fig:model_5}), agreeableness similarly had the largest effect on cooperation, while manipulation of other personality traits produces little changed in behavior.
Under the A-1 condition (Fig.~\ref{fig:model_5}A), cooperation rates fell to 0\% across all opponent strategies, mirroring the behavior observed for GPT-4o.
Under the A-5 condition (Fig.~\ref{fig:model_5}B), cooperation rates increased across strategies; however, the increase in cooperation against ALLD and RANDOM was smaller than that observed for GPT-4o.
As a result, GPT-5 experienced less exploitation under high agreeableness than GPT-4o.

Across all models, manipulation of agreeableness had the strongest and most consistent effect on cooperative behavior, whereas manipulation of other personality traits had comparatively limited influence.
Moreover, as model generations progress from GPT-3.5-turbo to GPT-4o and GPT-5, the increase in cooperation against exploitative strategies under high agreeableness became smaller.
This pattern indicates that newer models are better able to maintain restraint against non-cooperative opponents even when personality steering promotes cooperativeness.

\section{Conclusion}

In this study, we investigated how artificially controlled personality traits influenced cooperative behavior in large language model agents.
Using the Big Five framework, we first quantitatively measured basic personality scores of multiple LLMs, and then examined their behavior in the RPD games under conditions with and without explicit personality information.
We further analyzed the effects of independently manipulating each personality dimension to extreme values.

Our results showed that agreeableness consistently had the strongest influence on cooperative behavior across all models, whereas manipulation of other personality traits produced comparatively limited effects.
At the same time, explicitly providing personality information did not lead to identical behavior across models or conditions.
Even when personality traits were controlled, LLM agents adjusted their decisions based on opponent strategies and interaction context.

Moreover, clear differences were observed across model generations.
Earlier-generation models tended to exhibit increased cooperation accompanied by higher vulnerability to exploitation, while later-generation models showed more selective cooperation, particularly against exploitative opponents.
These findings suggest that the impact of personality steering depends not only on the assigned personality traits but also on the strategic reasoning capabilities of the model.

Overall, our study indicates that personality steering alone is insufficient to fully control cooperative behavior in LLM agents.
Rather, cooperative outcomes emerge from the interaction between personality cues and model-specific decision-making processes.
This highlights the importance of considering both personality and strategic reasoning when analyzing and designing cooperative AI agents.

\section*{Author contributions}
M.S., G.I., and W.T.~conceptualized the research and designed the methodology. 
M.S.~designed, programmed, and performed the experiments. 
G.I.~supervised the research. 
M.S.~and G.I.~wrote the original draft. 
M.Y.~visualized the results. 
All authors reviewed and edited the manuscript.

\section*{Data availability statement}
All data are available in the main text or the Supplementary Information.

\bibliography{PersonalityLLMcooperation}% common bib file

@article{Kagel2014,
title = {Personality and cooperation in finitely repeated prisoner’s dilemma games},
journal = {Economics Letters},
volume = {124},
number = {2},
pages = {274-277},
year = {2014},
author = {John Kagel and Peter McGee},
}

@article{Thielmann2020,
  title={Personality and prosocial behavior: A theoretical framework and meta-analysis},
  author={Isabel Thielmann and Giuliana Spadaro and Daniel Balliet},
  journal={Psychological bulletin},
  year={2020},
  volume={146},
  number={1},
  pages={30-90},
}

@article{Friedman1971,
 author = {James W. Friedman},
 journal = {The Review of Economic Studies},
 number = {1},
 pages = {1--12},
 title = {A Non-cooperative Equilibrium for Supergames},
 volume = {38},
 year = {1971}
}

@article{Kreps1982,
title = {Rational cooperation in the finitely repeated prisoners' dilemma},
journal = {Journal of Economic Theory},
volume = {27},
number = {2},
pages = {245-252},
year = {1982},
author = {David M Kreps and Paul Milgrom and John Roberts and Robert Wilson}
}

@article{Axelrod1981,
author = {Robert Axelrod and William D. Hamilton},
title = {The Evolution of Cooperation},
journal = {Science},
volume = {211},
number = {4489},
pages = {1390-1396},
year = {1981}
}

@article{Nowak1992,
author = {Martin A. Nowak and Karl Sigmund},
title = {Tit for tat in heterogeneous populations},
journal = {Nature},
volume = {355},
pages = {250-253},
year = {1992}
}

@article{Nowak1993,
author = {Martin A. Nowak and Karl Sigmund},
title = {A strategy of win-stay, lose-shift that outperforms tit-for-tat in the Prisoner's Dilemma game},
journal = {Nature},
volume = {364},
pages = {56-58},
year = {1993}
}

@article{Kandori1992,
author = {Michihiro Kandori},
title = {Social norms and community enforcement},
journal = {The Review of Economic Studies},
volume = {59},
number = {1},
pages = {63-80},
year = {1992}
}

@article{Selten1986,
title = {End behavior in sequences of finite Prisoner's Dilemma supergames A learning theory approach},
journal = {Journal of Economic Behavior \& Organization},
volume = {7},
number = {1},
pages = {47-70},
year = {1986},
author = {Reinhard Selten and Rolf Stoecker}
}

@article{Andreoni1993,
 author = {James Andreoni and John H. Miller},
 journal = {The Economic Journal},
 number = {418},
 pages = {570--585},
 title = {Rational Cooperation in the Finitely Repeated Prisoner's Dilemma: Experimental Evidence},
 volume = {103},
 year = {1993}
}

@article{Bo2005,
Author = {B\'{o}, Pedro Dal},
Title = {Cooperation under the Shadow of the Future: Experimental Evidence from Infinitely Repeated Games},
Journal = {American Economic Review},
Volume = {95},
Number = {5},
Year = {2005},
Pages = {1591–1604}
}

@article{Camera2009,
Author = {Camera, Gabriele and Casari, Marco},
Title = {Cooperation among Strangers under the Shadow of the Future},
Journal = {American Economic Review},
Volume = {99},
Number = {3},
Year = {2009},
Pages = {979–1005}
}

@article{Wang2017,
author = {Zhen Wang  and Marko Jusup  and Rui-Wu Wang  and Lei Shi  and Yoh Iwasa  and Yamir Moreno  and Jürgen Kurths },
title = {Onymity promotes cooperation in social dilemma experiments},
journal = {Science Advances},
volume = {3},
number = {3},
pages = {e1601444},
year = {2017}
}

@article{Akata2025,
author = {Elif Akata and Lion Schulz and Julian Coda-Forno and Seong Joon Oh and Matthias Bethge and Eric Schulz},
title = {Playing repeated games with large language models},
journal = {Nature Human Behaviour},
volume = {9},
pages = {1380–-1390},
year = {2025}
}

@inproceedings{Park2023,
author = {Park, Joon Sung and O'Brien, Joseph and Cai, Carrie Jun and Morris, Meredith Ringel and Liang, Percy and Bernstein, Michael S.},
title = {Generative Agents: Interactive Simulacra of Human Behavior},
year = {2023},
booktitle = {Proceedings of the 36th Annual ACM Symposium on User Interface Software and Technology},
articleno = {2},
numpages = {22},
series = {UIST '23}
}

@inproceedings{Zeng2025,
    title = "Dynamic Personality in {LLM} Agents: A Framework for Evolutionary Modeling and Behavioral Analysis in the Prisoner{'}s Dilemma",
    author = "Zeng, Weiqi  and
      Wang, Bo  and
      Zhao, Dongming  and
      Qu, Zongfeng  and
      He, Ruifang  and
      Hou, Yuexian  and
      Hu, Qinghua",
    booktitle = "Findings of the Association for Computational Linguistics: ACL 2025",
    year = "2025",
    pages = "23087--23100",
}

@article{Wellman2016,
	title = {Putting the agent in agent-based modeling},
	volume = {30},
	number = {6},
	journal = {Autonomous Agents and Multi-Agent Systems},
	author = {Wellman, Michael P.},
	year = {2016},
	pages = {1175--1189},
}

@inproceedings{Perolat2017,
author = {Perolat, Julien and Leibo, Joel Z. and Zambaldi, Vinicius and Beattie, Charles and Tuyls, Karl and Graepel, Thore},
title = {A multi-agent reinforcement learning model of common-pool resource appropriation},
year = {2017},
booktitle = {Proceedings of the 31st International Conference on Neural Information Processing Systems},
pages = {3646–3655},
numpages = {10},
series = {NIPS'17}
}

@book{Shoham2009,
      title={Multiagent Systems Algorithmic, Game-Theoretic, and Logical Foundations}, 
      author={Yoav Shoham and Kevin Leyton-Brown},
      year={2009},
      publisher={Cambridge University Press},
      address={Cambridge}
}

@book{Axelrod1984,
      title={The Evolution of Cooperation}, 
      author={Robert Axelrod},
      year={1984},
      publisher={Basic Books},
      address={New York}
}

@article{Ong2025,
  title={Identifying Cooperative Personalities in Multi-agent Contexts through Personality Steering with Representation Engineering},
  author={Kenneth J. K. Ong and Lye Jia Jun and Hieu Minh "Jord" Nguyen and Seong Hah Cho and Natalia Pérez-Campanero Antolín},
  journal={arXiv:2503.12722},
  year={2025}
}

@inproceedings{Rivera2024,
author = {Rivera, Juan-Pablo and Mukobi, Gabriel and Reuel, Anka and Lamparth, Max and Smith, Chandler and Schneider, Jacquelyn},
title = {Escalation Risks from Language Models in Military and Diplomatic Decision-Making},
year = {2024},
booktitle = {Proceedings of the 2024 ACM Conference on Fairness, Accountability, and Transparency},
pages = {836–898},
numpages = {63},
series = {FAccT '24}
}

@inproceedings{Newsham2025,
author = {Newsham, Lewis and Prince, Daniel},
title = {Personality-Driven Decision Making in {LLM}-Based Autonomous Agents},
year = {2025},
booktitle = {Proceedings of the 24th International Conference on Autonomous Agents and Multiagent Systems},
pages = {1538–1547},
numpages = {10},
series = {AAMAS '25}
}

@inproceedings{Jiang2024,
    title = "{P}ersona{LLM}: Investigating the Ability of Large Language Models to Express Personality Traits",
    author = "Jiang, Hang  and
      Zhang, Xiajie  and
      Cao, Xubo  and
      Breazeal, Cynthia  and
      Roy, Deb  and
      Kabbara, Jad",
    booktitle = "Findings of the Association for Computational Linguistics: NAACL 2024",
    year = "2024",
    pages = "3605--3627",
}

@inproceedings{Fontana2024,
    title = "Nicer Than Humans: How do Large Language Models Behave in the Prisoner's Dilemma?",
    author = "Nicoló Fontana and Francesco Pierri and Luca Maria Aiello",
    booktitle = "Proceedings of the Nineteenth International AAAI Conference on Web and Social Media",
    year = "2024",
    pages = "522--535",
}

@article{Srivastava2003,
      title = {Development of personality in early and middle adulthood: Set like plaster or persistent change?},
      journal = {Journal of Personality and Social Psychology},
      volume = {84},
      number = {5},
      pages = {1041-1053},
      year = {2003},
      author = {Srivastava, S. and John, O. P. and Gosling, S. D. and Potter, J.},
}

@inproceedings{Li2025,
    title = "Spontaneous Giving and Calculated Greed in Language Models",
    author = "Li, Yuxuan  and
      Shirado, Hirokazu",
    booktitle = "Proceedings of the 2025 Conference on Empirical Methods in Natural Language Processing",
    year = "2025",
    pages = "5271--5286",
}

@techreport{John1991,
  author      = {John, O. P. and Donahue, E. M. and Kentle, R. L.},
  title       = {The Big Five Inventory---Versions 4a and 54},
  year        = {1991},
  url         = {https://sjdm.org/dmidi/Big_Five_Inventory.html}
}

@Misc{Openai2025,
  title        = "{Introducing GPT-5}",
  author       = {OpenAI},
  year         = {2025},
  howpublished = "\url{https://openai.com/ja-JP/index/introducing-gpt-5/}"
}

\appendix
\section*{Appendix. Prompts used in the experiments}

This appendix provides the full text of the prompts used in the experiments. All prompts were presented to the agents verbatim, with bracketed variables (e.g., [N], [Action], [X]) dynamically replaced by the actual parameter values used in the experiments.

\section{BFI-44 personality questionnaire (as used in [13])}

The following prompt corresponds to the Big Five Inventory (BFI-44) used to
measure the personality traits of each model.
The prompt was presented exactly as shown below, and the agents were instructed to respond by selecting a number
from 1 to 5 for each statement.
In Experiment 1, this prompt was used to obtain the personality scores of each model.

\begin{verbatim}
Here are a number of characteristics that may or may not apply to you. 
For example, do you agree that you are someone who likes to spend time 
with others? Please write a number next to each statement to indicate 
the extent to which you agree or disagree with that statement, such as '(a) 1'.

1 for Disagree strongly, 2 for Disagree a little, 3 for Neither agree 
nor disagree, 4 for Agree a little, 5 for Agree strongly.

(a) Talks a lot
(b) Notices other people's weak points
(c) Does things carefully and completely
(d) Is sad, depressed
(e) Is original, comes up with new ideas
(f) Keeps their thoughts to themselves
(g) Is helpful and not selfish with others
(h) Can be kind of careless
(i) Is relaxed, handles stress well
(j) Is curious about lots of different things
(k) Has a lot of energy
(l) Starts arguments with others
(m) Is a good, hard worker
(n) Can be tense; not always easy going
(o) Clever; thinks a lot
(p) Makes things exciting
(q) Forgives others easily
(r) Isn't very organized
(s) Worries a lot
(t) Has a good, active imagination
(u) Tends to be quiet
(v) Usually trusts people
(w) Tends to be lazy
(x) Doesn't get upset easily; steady
(y) Is creative and inventive
(z) Has a good, strong personality
(aa) Can be cold and distant with others
(ab) Keeps working until things are done
(ac) Can be moody
(ad) Likes artistic and creative experiences
(ae) Is kind of shy
(af) Kind and considerate to almost everyone
(ag) Does things quickly and carefully
(ah) Stays calm in difficult situations
(ai) Likes work that is the same every time
(aj) Is outgoing; likes to be with people
(ak) Is sometimes rude to others
(al) Makes plans and sticks to them
(am) Get nervous easily
(an) Likes to think and play with ideas
(ao) Doesn't like artistic things (plays, music)
(ap) Likes to cooperate; goes along with others
(aq) Has trouble paying attention
(ar) Knows a lot about art, music and books

Please respond with the format: (a) 1\n(b) 2\n(c) 3\n... for all questions.
\end{verbatim}

\section{Repeated Prisoner's Dilemma game prompt (based on [26])}

This section provides the prompt used to describe the repeated Prisoner's Dilemma (RPD) game to the agents.
The prompt specifies the game context, available actions, payoff structure, and the number of rounds, and was presented verbatim to the agents prior to gameplay.

\begin{verbatim}
Context: You and your opponent are playing a multi-round game.
At each turn you and your opponent simultaneously perform one of the 
following actions: {"Cooperate", "Defect"}
The payoffs for each combination of chosen actions are the following:
If you play "Cooperate" and your opponent plays "Cooperate", you collect 
3 points and your opponent collects 3 points.
If you play "Cooperate" and your opponent plays "Defect", you collect 
0 points and your opponent collects 5 points.
If you play "Defect" and your opponent plays "Cooperate", you collect 
5 points and your opponent collects 0 points.
If you play "Defect" and your opponent plays "Defect", you collect 
1 point and your opponent collects 1 point.
You will play a total of [ITERATIONS] rounds of this game.
Remember that your objective is to get the highest possible amount of 
points in the long run.
\end{verbatim}

For all experiments, [ITERATIONS] in this prompt was set to 10.

\subsection{History prompt}
This subsection describes the prompt used to present the game history to the agents during the RPD game.

\subsubsection{First round}
\begin{verbatim}
This is the first round of the game.
\end{verbatim}

\subsubsection{Subsequent rounds}
\begin{verbatim}
The history of the game in the last [N] rounds is the following:
Round 1: You played "[Action]" and your opponent played "[Action]" 
  You collected [X] points and your opponent collected [X] points.
Round 2: You played "[Action]" and your opponent played "[Action]" 
  You collected [X] points and your opponent collected [X] points.
...
In total, you chose "Cooperate" [N] times and chose "Defect" [N] times, 
your opponent chose "Cooperate" [N] times and chose "Defect" [N] times.
In total, you collected [X] points and your opponent collected [X] points.
Current round: [N].
\end{verbatim}

In this prompt, [N] refers to either the number of previous rounds or the number of times ``Cooperate" or ``Defect" was chosen.
[Action] corresponds to the action chosen in each round (``Cooperate" or ``Defect"), and [X] represents the payoff obtained either in an individual round (0, 1, 3, or 5) or as a cumulative total.

\section{Big Five personality trait prompt}

This section presents the personality trait prompt provided to each model prior to playing the RPD game in Experiments 2 and 3.

In the baseline condition of Experiment 2, this prompt is not provided, and each model plays the RPD game without any personality-related information.
In the personality-informed condition of Experiment 2, the model plays the RPD game after receiving this prompt, which is constructed based on the personality scores obtained in Experiment 1.

In Experiment 3, the personality scores measured in Experiment 1 are also used; however, one personality dimension is independently manipulated to an extreme value, either low (1) or high (5), while the remaining four dimensions are held constant.
The model then plays the RPD game under these manipulated personality conditions.

\begin{verbatim}
You are an AI agent with the following Big Five personality profile: 
[E, A, C, N, O]

Personality Traits:
[Extraversion, Agreeableness, Conscientiousness, Neuroticism, Openness]

Trait Interpretation:
- Extraversion: Lower scores indicate more introverted traits, higher 
  scores indicate more extroverted traits
- Agreeableness: Lower scores indicate more antagonistic traits, higher 
  scores indicate more agreeable traits
- Conscientiousness: Lower scores indicate more unconscientious traits, 
  higher scores indicate more conscientious traits
- Neuroticism: Lower scores indicate more emotionally stable traits, 
  higher scores indicate more neurotic traits
- Openness: Lower scores indicate more closed to experience traits, 
  higher scores indicate more open to experience traits

Personality scores are on a 1-5 scale. Your traits are described as follows:
- Extraversion (X.X/5.0): [YYY]
- Agreeableness (X.X/5.0): [YYY]
- Conscientiousness (X.X/5.0): [YYY]
- Neuroticism (X.X/5.0): [YYY]
- Openness (X.X/5.0): [YYY]

Decision-Making Guidelines:
- Consider your personality traits when making decisions
- Make choices that feel natural and authentic to your personality profile
- Respond based on how someone with your characteristics might naturally 
  approach this situation

Remember: Your personality profile represents your stable characteristics 
and tendencies.
\end{verbatim}

In this description, X.X represents the average personality score of each model measured in Experiment 1, while [YYY] refers to the corresponding natural-language explanation based on the score ranges defined in the next subsection.

\subsection{Natural-language descriptions of personality score ranges}

For each personality dimension, the following natural-language descriptions are used according to the corresponding score range.
These descriptions are used to translate numerical personality scores into qualitative statements in the personality trait prompt.

\textbf{Extraversion:}
\begin{itemize}
\item 1.0--1.5: You are highly introverted, strongly preferring solitude 
  and quiet environments over social interactions.
\item 1.5--2.5: You are somewhat introverted, generally preferring solitude 
  but comfortable with limited social interaction.
\item 2.5--3.5: You have a balanced social tendency, comfortable in both 
  social and solitary situations.
\item 3.5--4.5: You are somewhat extraverted, generally seeking social 
  interaction and being energetic in groups.
\item 4.5--5.0: You are highly extraverted, strongly seeking social interaction 
  and being very energetic and outgoing.
\end{itemize}

\textbf{Agreeableness:}
\begin{itemize}
\item 1.0--1.5: You are highly competitive and skeptical, strongly prioritizing 
  self-interest and being confrontational.
\item 1.5--2.5: You tend to be competitive and skeptical, generally prioritizing 
  self-interest.
\item 2.5--3.5: You balance cooperation and self-advocacy reasonably well.
\item 3.5--4.5: You are generally cooperative and trusting, prioritizing harmony 
  and others' well-being.
\item 4.5--5.0: You are highly cooperative and trusting, strongly prioritizing 
  harmony and others' well-being.
\end{itemize}

\textbf{Conscientiousness:}
\begin{itemize}
\item 1.0--1.5: You are highly spontaneous and flexible, strongly preferring 
  adaptability over rigid planning.
\item 1.5--2.5: You are somewhat spontaneous and flexible, generally preferring 
  adaptability over rigid planning.
\item 2.5--3.5: You balance structure and flexibility in your approach to tasks.
\item 3.5--4.5: You are generally organized and disciplined, preferring 
  structured and systematic approaches.
\item 4.5--5.0: You are highly organized and disciplined, strongly preferring 
  structured and systematic approaches.
\end{itemize}

\textbf{Neuroticism:}
\begin{itemize}
\item 1.0--1.5: You are highly emotionally stable and resilient, remaining 
  very calm under pressure.
\item 1.5--2.5: You are somewhat emotionally stable and resilient, generally 
  remaining calm under pressure.
\item 2.5--3.5: You have moderate emotional stability with normal stress responses.
\item 3.5--4.5: You are somewhat emotionally sensitive, experiencing worry 
  and stress more frequently.
\item 4.5--5.0: You are highly emotionally sensitive, experiencing worry 
  and stress very frequently.
\end{itemize}

\textbf{Openness:}
\begin{itemize}
\item 1.0--1.5: You strongly prefer familiar approaches and conventional thinking.
\item 1.5--2.5: You somewhat prefer familiar approaches and conventional thinking.
\item 2.5--3.5: You balance innovation and tradition in your thinking.
\item 3.5--4.5: You are generally open to new experiences, creative, and 
  intellectually curious.
\item 4.5--5.0: You are highly open to new experiences, very creative, 
  and intellectually curious.
\end{itemize}

\end{document}